\documentclass[conference]{IEEEtran}  

\IEEEoverridecommandlockouts                              

\usepackage{xcolor}
\usepackage{textcomp}
\usepackage{graphicx}
\usepackage{textcomp}
\usepackage{hyperref}
\newcommand{\TEACar}{\textsc{TEACar}}

\title{\LARGE \bf
TEACar: An Open-Source Autonomous Driving Platform
}

\author{Zhongzheng Zhang$^{1}$, Maxwell Ruyle$^{1}$, Andrew Kappes$^{1}$,\\ Tyler Ruble$^{1}$, William Shaoul$^{1}$, Dana Moreno$^{1}$, Jack Penn$^{1}$ and Ivan Ruchkin$^{1}$
\thanks{$^{1}$Trustworthy Engineered Autonomy (TEA) Lab, Department of Electrical and Computer Engineering, University of Florida,
{\tt\small renzhongzh.zhang@ufl.edu, mruyle@ufl.edu, andrew.kappes@ufl.edu, tyler.ruble@ufl.edu, wshaoul@ufl.edu, d.moreno@ufl.edu, j.penn@ufl.edu, iruchkin@ece.ufl.edu}
}%

}

\begin{document}

\maketitle
\thispagestyle{empty}
\pagestyle{empty}

\begin{abstract}
\looseness=-1
Intelligent Transportation Systems (ITS) increasingly rely on vision-based perception and learning-based control, necessitating experimental platforms that support realistic hardware-in-the-loop validation. Small-scale platforms for autonomous racing offer a practical path to hardware validation --- but often suffer from limited modularity, high integration complexity, or restricted extensibility. This paper presents \TEACar, a 1/14- to 1/16-scale autonomous driving platform designed with modular mechanical architecture, hardware abstraction, and ROS~2-based software. The system adopts a four-layer deck structure that physically decouples sensing, computation, actuation, and power subsystems, improving structural rigidity while simplifying reconfiguration. 
We constructed and comprehensively evaluated the prototype of \TEACar. Its mechanical stability, structural characteristics, and software performance were quantified based on three CNN-based steering controllers. Inference latency, power consumption, and system operating time were measured to evaluate computational capability and robustness. Our experiments demonstrated that \TEACar~
offers a scalable, modular, and cost-effective testbed for ITS research, education, and development. Our \href{https://github.com/Trustworthy-Engineered-Autonomy-Lab/TEACar-Open-Source-Autonomous-Driving-Platform.git}{project repository} is available on GitHub.
\end{abstract}

\begin{IEEEkeywords}
Autonomous driving, autonomous racing, hardware platform, modularity 
\end{IEEEkeywords}

\section{Introduction}

Intelligent Transportation Systems (ITS) integrate computational intelligence with dynamic physical environments. Research efforts span perception, decision-making, control, communication, and system-level optimization~\cite{ASLAM2026143}. In particular, vision-based tasks constitute a major component of modern ITS research~\cite{s25082611,doi:10.1177/09544070231203059,zhang2024researchapplicationcomputervision}, enabling capabilities such as lane detection, obstacle recognition, localization, and learning-based control. The widespread adoption of camera sensors and advances in machine learning have further reinforced the importance of vision-driven approaches in ITS.

\looseness=-1
Experimental hardware-in-the-loop evaluation is essential to ITS research and education: algorithms and models must be validated under realistic operating conditions. Although simulation environments, such as Gazebo, provide convenient and cost-effective early-stage development and testing, they cannot fully capture the complexity and uncertainty inherent in physical systems~\cite{hu2023simulationhelpsautonomousdrivinga,aljalbout2025realitygaproboticschallenges}. Consequently, physical experimental platforms remain crucial for rigorous validation, performance evaluation, and system-level analysis. However, constructing and maintaining full-scale vehicles for experimental verification is often time-consuming and financially prohibitive, often requiring substantial infrastructure, safety considerations, and operational resources. 

\looseness=-1
In contrast, small-scale autonomous vehicle platforms offer a practical and scalable alternative. Despite their reduced physical dimensions, these platforms preserve the essential sensing-computation-control feedback loop found in full-scale systems. They are sufficient to validate perception, planning, and control algorithms. Two small-scale experimental platforms stand out. Among the most widely adopted is \textit{RoboRacer} (formerly F1/10 or F1TENTH)~\cite{okelly2019f110opensourceautonomouscyberphysical}, a 1/10-scale autonomous vehicle testbed designed to reduce the burden on researchers and educators by providing a standardized and reproducible experimental platform~\cite{betz2022teachingautonomoussystemshandson}. However, its large physical scale and high cost reduce its suitability for vision-based research tasks, while the lack of modular design limits flexibility and complicates system reconfiguration. \textit{Donkeycar}~\cite{donkeycar_docs} represents another popular small-scale platform that is primarily designed for vision-based tasks and offers a high degree of customization, making it particularly attractive for learning-based research and educational use. However, its non-modular software design and reliance on separate power sources for computation and actuation increase system integration complexity and limit extensibility.

\begin{figure*}[tbh]
  \centering\includegraphics{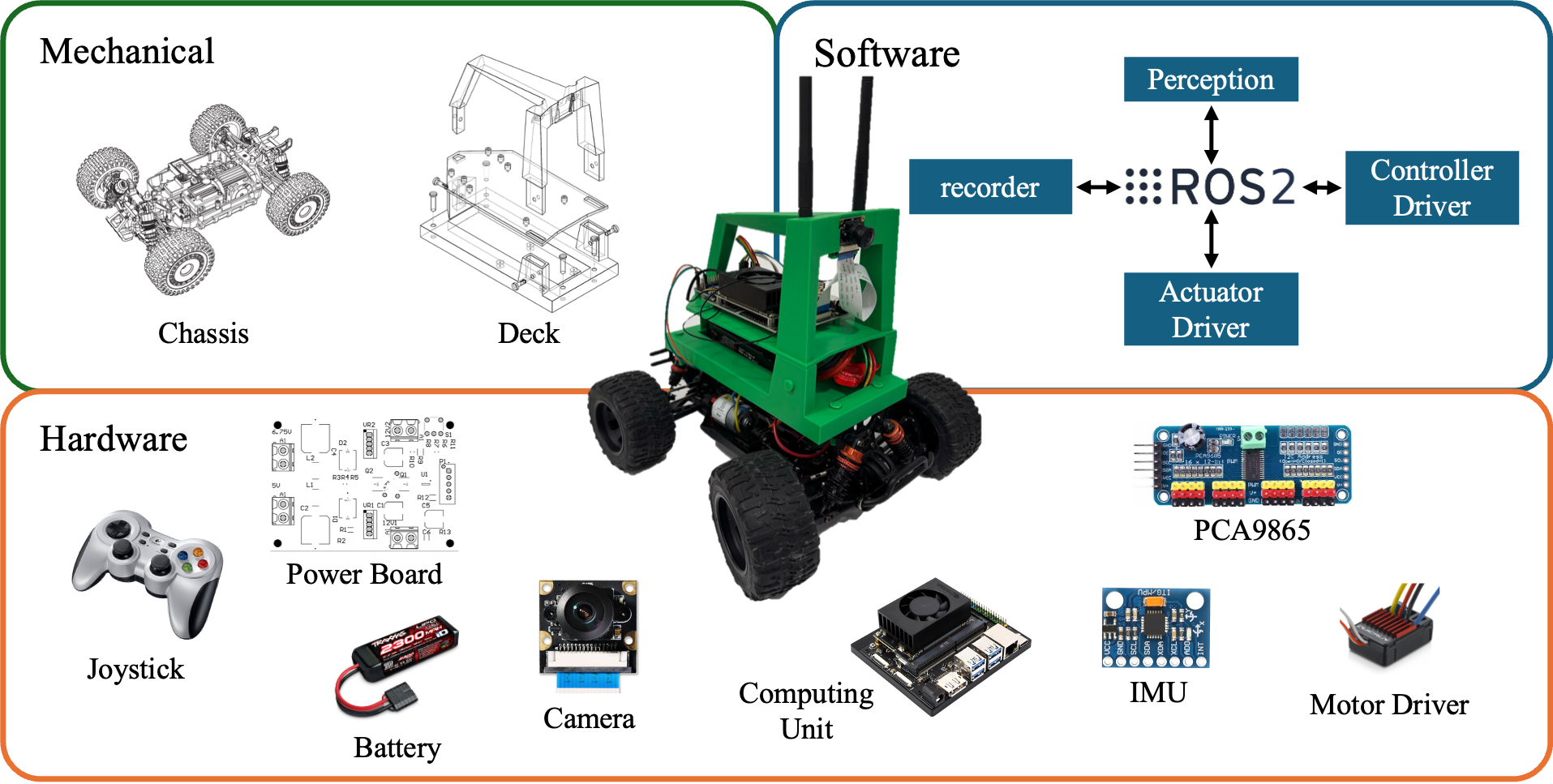}
  \caption{Overview of the \TEACar~platform.}
  \label{fig:system_overview}
\end{figure*}

To overcome these challenges, we developed \TEACar, a 1/14- to 1/16-scale autonomous racing platform explicitly designed around modularity and hardware abstraction. The platform is intended to support a wide range of experimental configurations. At the mechanical level, the platform adopts a four-layer deck architecture that physically decouples key subsystems, improving mechanical stability while simplifying assembly and component replacement. On the software side, the platform offers a standardized, extensible middleware layer for sensing, communication, and control via the Robot Operating System (ROS) 2~\cite{doi:10.1126/scirobotics.abm6074}. The hardware ensures safe and reliable system operation with a dedicated power distribution board, designed to support onboard components with varying voltage requirements.

To illustrate the use of our platform, we constructed a \TEACar~prototype and conducted its comprehensive evaluation, including the mechanical characteristics and structural rigidity of the design. In addition, software performance was evaluated by deploying three CNN-based controllers that generated steering commands from camera images. For each neural controller, we measured the inference latency, power consumption, and system operating time. The experimental results demonstrate that the platform provides a mechanically stable structure while maintaining sufficient computational performance for learning-based control tasks. These findings indicate that the proposed design is well-suited for autonomous driving research, education, and development.

\looseness=-1
The remainder of this paper is organized as follows. 
Section~\ref{sec:related_work} reviews related small-scale platforms for autonomous racing. 
Section~\ref{sec:system_overview} provides an overview of the system and its design goals. 
Sections~\ref{sec:mechanical_design}, \ref{sec:hardware_design}, and \ref{sec:software_desgin} describe the mechanical, hardware, and software design of \TEACar, respectively. 
Section~\ref{sec:evaluation} presents the experimental evaluation results, and Section~\ref{sec:future_work} concludes the paper and discusses future works.

\section{Related Work}\label{sec:related_work}

\looseness=-1
\noindent
\textbf{RoboRacer}. RoboRacer (formerly F1TENTH or F1/10) is a widely adopted 1/10-scale autonomous vehicle research platform originally introduced by the University of Pennsylvania~\cite{okelly2019f110opensourceautonomouscyberphysical}. Since its initial development, the platform has evolved into a large open research community, serving as a common experimental testbed for autonomous driving studies~\cite{okelly2019f110opensourceautonomouscyberphysical}. The system integrates a high-performance embedded computing unit, typically based on NVIDIA\textregistered~Jetson devices, together with multiple onboard sensors, including LiDAR and depth cameras~\cite{okelly2019f110opensourceautonomouscyberphysical}. This hardware configuration enables the execution of computationally intensive perception, planning, and control algorithms, supporting a broad range of research directions such as state estimation, mapping, learning-based control, and real-time decision-making~\cite{pmlr-v123-o-kelly20a,evans2024unifyingf1tenthautonomousracing}.

\looseness=-1
The RoboRacer platform benefits from the computational capability of Jetson-class embedded processors, a well-engineered power distribution system that supports stable operation of onboard components, and a ROS-based software ecosystem that facilitates algorithm development and system integration. These design choices inspired aspects of our work. However, it has several drawbacks. First, the platform is tightly coupled to the Traxxas\textregistered~1/10-scale chassis, limiting flexibility for mechanical redesign and hardware customization. Second, its relatively large physical footprint and turning radius constrain deployment in space-limited laboratory environments. Third, the system cost and hardware configuration may be excessive for vision-based research tasks.

\noindent
\textbf{Donkeycar}. Donkeycar is a low-cost autonomous vehicle research and educational platform developed as an open-source framework for scaled self-driving experimentation~\cite{donkeycar_docs}. Originally designed to emphasize accessibility and rapid prototyping, the platform has grown into a community-driven ecosystem, serving as a practical testbed for learning-based autonomous driving studies~\cite{donkeycar_docs}. The system is typically built upon lightweight embedded computing units, such as Raspberry Pi\textregistered~devices, combined with consumer-grade sensors, including front-facing cameras and optional inertial units~\cite{donkeycar_docs}. This hardware architecture supports real-time data collection and onboard inference, enabling a variety of research directions such as imitation learning, reinforcement learning, end-to-end control, and sim-to-real policy transfer~\cite{TRENTSIOS2022287,mahajan2024quantifyingsim2realgapgps}.

The small physical scale of the platform makes it well-suited for laboratory environments and rapid experimental iteration, which motivated our adoption of a similar scale. However, the platform design introduces several practical constraints. First, the use of separate power sources for computation and actuation increases hardware complexity and integration overhead. Second, the predominantly Python-based custom software stack~\cite{donkeycar_github}, while highly accessible, may limit real-time performance for computationally intensive workloads and reduce system-level extensibility. Third, although the platform can be adapted to multiple chassis configurations, doing so typically requires replacement or substantial redesign of the 3D-printed structural components, limiting rapid mechanical reconfiguration.

\section{System Overview}\label{sec:system_overview}
The platform consists of three systems: the mechanical system, the hardware system, and the software system. The whole system architecture is shown in Fig.~\ref{fig:system_overview}

We believe that the next-generation autonomous racing platform should satisfy three system-level requirements:

\noindent
\textbf{Compactness}: The system must maintain a space-efficient design, considering both the physical dimensions of the vehicle and the operating space required for constructing and running the experimental track.

\noindent
\textbf{Modularity}: The system architecture must support flexible integration, replacement, and extension of hardware and software components. This will enable users to plug in their techniques and algorithms as components.


\noindent
\textbf{Reliability}: The platform should ensure normal operation under varying computational and electrical workloads.

The next three sections describe how the mechanical, hardware, and software systems fulfill these requirements.

\section{Mechanical Design}\label{sec:mechanical_design}
\looseness=-1
The objective of mechanical design is to create a modular structure that supports autonomous hardware while staying adaptable to different small-scale RC chassis.
The mechanical system of the \TEACar~platform consists of two primary components: the chassis of an RC vehicle and the deck. 

\subsection{Chassis}

The chassis serves as the base for mobility and is responsible for steering, propulsion, and suspension. In small-scale autonomous racing research, common platforms include commercial 1/14- and 1/16-scale RC vehicles, such as those produced by Traxxas\textregistered~and ARRMA\textregistered. This range of scales was chosen to fit into constrained lab and classroom spaces, unlike the 1/10-scale chassis, which requires significantly larger tracks due to its turning radius. 

The available commercial RC chassis vary in wheelbase, mounting geometry, suspension layout, and available packaging volume. 
To avoid dependence on a single vehicle configuration (which may be unavailable to the users), the chassis is treated as a generalized mobility substrate, assuming it provides a rigid mounting region, sufficient payload capacity for the deck and onboard hardware, and steering and throttle actuation that accept external control signals. 


\subsection{Deck}

\begin{figure}[htbp]
  \centering
  \includegraphics{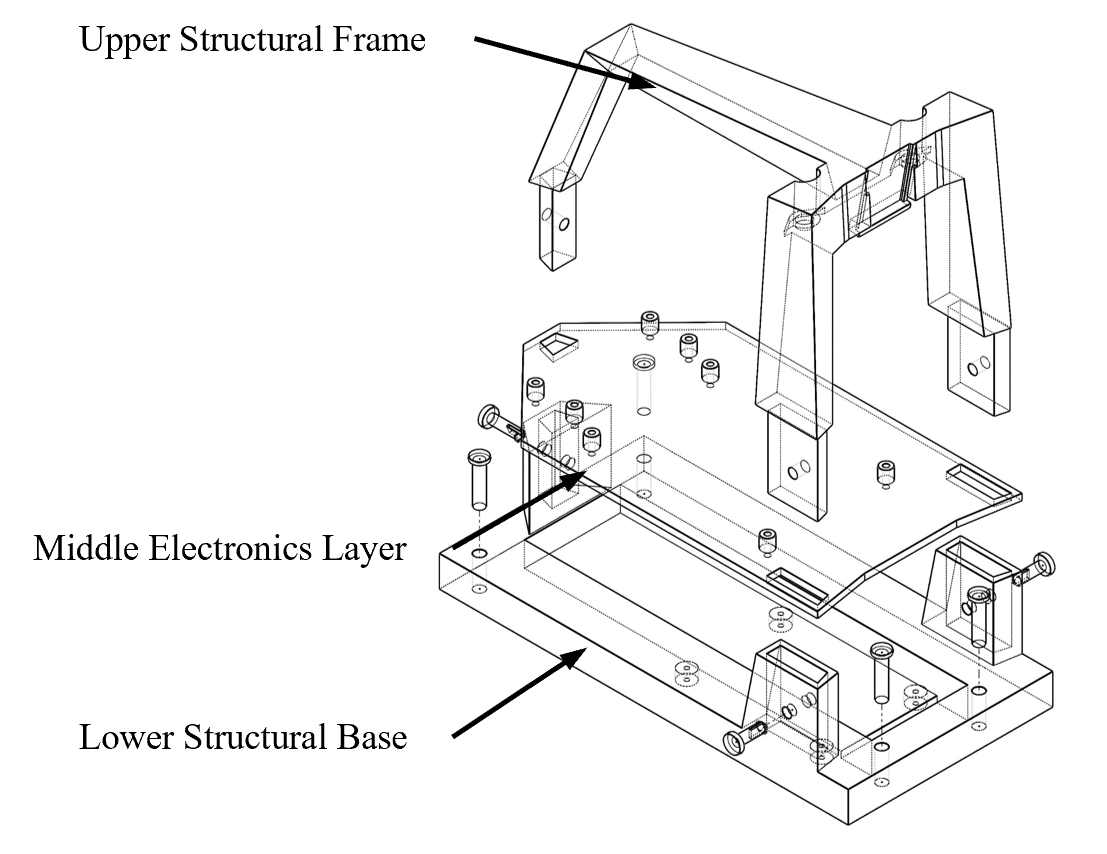}
  \vspace{-3mm}
  \caption{Overview of the \TEACar~deck.}
  \label{fig:deck_overview}
  \vspace{-4mm}
\end{figure}

The deck is a modular structural assembly that standardizes the physical packaging and mounting of computing hardware, power systems, and sensors independently of the RC chassis. Rather than adapting components to a specific chassis layout, the deck provides a unified structure that interfaces with multiple 1/14- and 1/16-scale RC platforms through a modular adapter layer that isolates chassis-specific geometry. The deck is organized into \textit{four functional layers}, as illustrated in Fig.~\ref{fig:deck_overview}.

\smallskip
\noindent
\textbf{Chassis Adapter Interface Layer}. This layer mechanically bridges the chassis and the deck and is interchangeable to match chassis-specific mounting geometries, including hole spacing, fastener diameter, vertical interface height, and local structural stiffness. The adapter defines the load transfer path between the vehicle suspension system and the deck and ensures consistent alignment of the platform relative to the chassis centerline. By isolating variations in wheelbase, mounting hole patterns, and deck elevation to this layer, the remaining deck geometry remains invariant across compatible 1/14- and 1/16-scale chassis.

\smallskip
\noindent
\textbf{Lower Structural Base}. The lower structural base interfaces directly with the adapter layer and supports primary structural loads. This layer houses the main battery pack and provides constrained battery retention under acceleration, braking, and lateral cornering. The base is designed to maintain a low center of gravity while allowing accessible battery insertion and removal without disturbing the upper layers. It also distributes reaction forces from the upper frame into the chassis through the adapter interface.

\looseness=-1
\smallskip
\noindent
\textbf{Middle Electronics Layer}. The middle layer provides a rigid mounting surface for the computing unit (e.g., NVIDIA\textregistered~Jetson modules), power distribution board, motor driver, actuator board, and IMU. The layer includes predefined mounting hole locations to secure computing and power modules while maintaining spacing between components for access and thermal clearance. The flat geometry simplifies installation and replacement of components while maintaining consistent alignment across configurations.

\smallskip
\noindent
\textbf{Upper Structural Frame}. The upper frame supports forward-facing sensors, including CSI or USB cameras, and it provides a mounting location for wireless communication antennas while providing mechanical protection during rollovers or handling impacts. The structure elevates sensors to improve the field-of-view and reduce occlusion from chassis components. In addition to impact shielding, the frame functions as a rigid grab structure for safe manual transport and recovery of the vehicle without applying load directly to sensitive electronics.


\looseness=-1
When transitioning to a new chassis, \textit{only} the chassis adapter interface layer requires redesign and replacement, while the remaining deck layers remain unchanged. This modular architecture localizes geometric variation to a single component, minimizes manufacturing overhead, shortens integration time, and preserves consistent hardware placement and center-of-gravity across different configurations.

\section{Hardware Design}\label{sec:hardware_design}
\begin{figure}[htbp]
  \centering
  \includegraphics{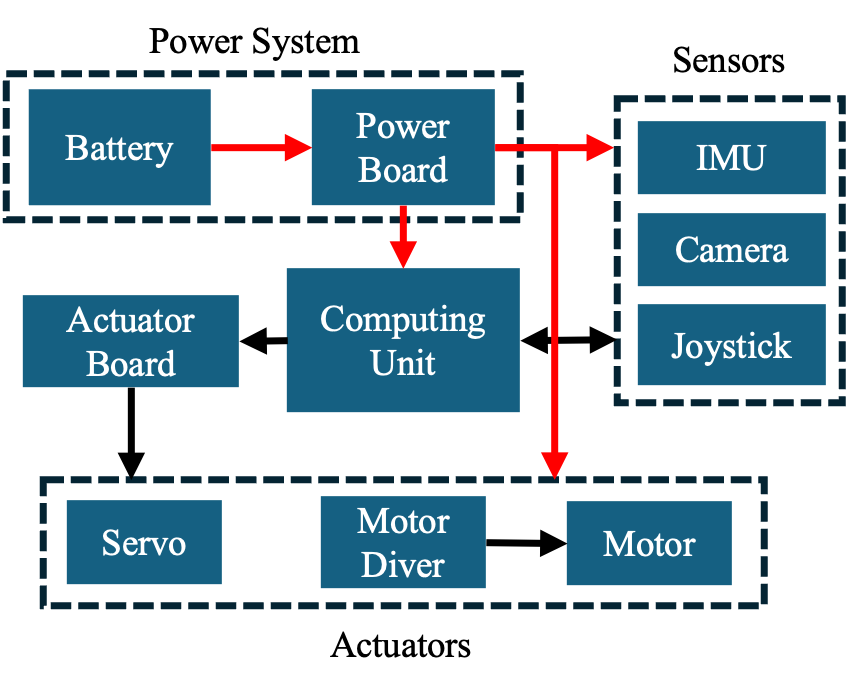}
  \vspace{-5mm}
  \caption{Hardware system architecture.}
  \label{fig:hardware_system}
  \vspace{-3mm}
\end{figure}

\looseness=-1
The hardware system, which forms the core of the platform, serves as the interface between the software stack and the mechanical components. The primary objective of this system is to perform signal exchange and power delivery, thereby ensuring reliable execution of sensing, computation, and actuation functions. As illustrated in Fig.~\ref{fig:hardware_system}, the system consists of a compute unit, an actuator board, a power system, sensors, and actuators.

\subsection{Computing Unit}
The computing unit, the brain of the car, is where software programs run. Its performance is vital for the whole system. Our platform supports multiple NVIDIA\textregistered~edge computing modules, including the \textit{Jetson Xavier NX series}, \textit{Jetson Orin NX series}, and \textit{Jetson Orin Nano\texttrademark~series}. These devices offer AI performance ranging from 21 TOPS to 100 TOPS, enabling the execution of a wide variety of AI workloads. Despite their size being comparable to a credit card, they provide substantial computational capability, supporting full-scale operating systems, and can be readily integrated into resource-constrained systems. 

\subsection{Power System}
The power system is responsible for providing a continuous power supply to all onboard components under external disturbances, including mechanical vibrations encountered during vehicle motion. This system consists of two primary elements: the battery and the power board.

\smallskip
\noindent
\textbf{Battery}. The battery serves as the energy source of the system. The platform supports both LiPo and NiMH battery configurations designed to satisfy the required operating voltage range of 9–15 V. In practice, this requirement can typically be achieved using a 3S LiPo pack or an equivalent NiMH pack configuration. LiPo batteries are generally preferred due to their higher energy density, higher nominal cell voltage, and lower internal resistance~\cite{KANG2014618}. These advantages can extend operating time and enhance acceleration performance.

\begin{figure}[htbp]
  \centering\includegraphics{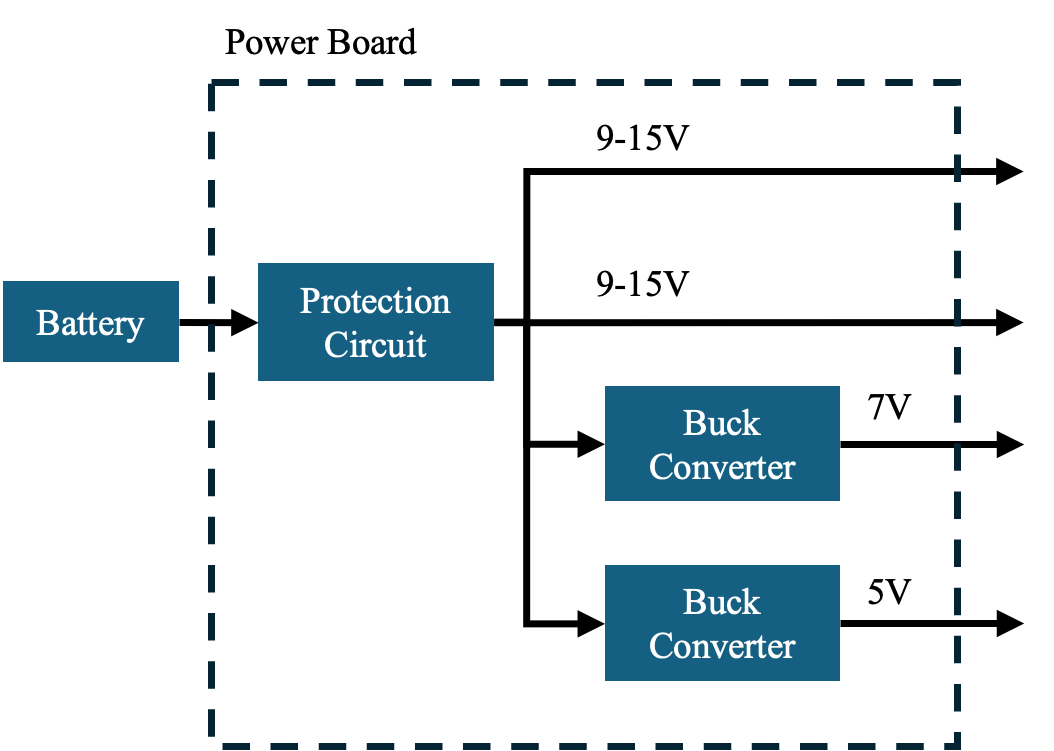}
  \vspace{-2mm}
  \caption{Power board architecture.}
  \label{fig:power_board}
  \vspace{-3mm}
\end{figure}

\smallskip
\noindent
\textbf{Power Board}. The power distribution board delivers power to onboard components with varying voltage requirements while preventing excessive battery discharge and protecting the system against abnormal voltage conditions. The computing unit, which requires a 9--15~V input, and the motor driver, designed for a 3S battery supply, are powered directly from the battery input. The servo motor is supplied through a regulated low-voltage rail generated by an integrated buck converter, ensuring operation below 7~V. In addition, the board provides a dedicated 5~V regulated output rail to power onboard sensors and other low-voltage peripherals. The board further incorporates a protection circuit that disables the outputs when the input voltage falls outside predefined safety limits. The functional architecture of the board is illustrated in Fig.~\ref{fig:power_board}.

\subsection{Sensors}
\looseness=-1
Sensors are necessary for driving tasks, such as line following and competitive racing. The platform supports three sensor modalities: cameras, joysticks, and inertial measurement units (IMUs).


\looseness=-1
For vision sensing, the platform accommodates both CSI and USB cameras with varying specifications. For inertial sensing, the platform is compatible with MPU-series\textregistered~IMUs, including MPU6050 and MPU9250. For manual operation, the platform supports a USB-connected wireless joystick.

\subsection{Actuators and Actuator Board}
\looseness=-1
Actuators serve as the interface between the electrical and mechanical systems, converting electrical signals into physical motion. Our platform supports up to 16 actuators controlled with Pulse Width Modulation (PWM), in which variations in pulse width correspond to position or velocity commands. However, typical onboard computing units do not natively provide a sufficient number of PWM channels capable of generating the precise timing signals required for reliable direct actuator control. To overcome this limitation, a PCA9685-based PWM controller board was integrated into the system. The board communicates with the computing unit via the I2C protocol, which is widely supported by embedded computing platforms, and provides up to 16 independent PWM output channels. Our platform requires two primary actuators: a motor and a servo.

\looseness=-1
\smallskip
\noindent
\textbf{Servo}.
The servo gets PWM, generates the steering motion, and operates within its specified supply voltage range of 5--7~V.

\looseness=-1
\smallskip
\noindent
\textbf{Motor}.
The motor needs an appropriate motor driver that interprets the PWM signal as a throttle command and operates within an input voltage range of 9--15~V. However, due to the variability in motor specifications across different RC chassis, it is impractical to design a universal motor driver that suits all configurations. To maintain platform adaptability, our design does not include a predefined motor or driver; instead, users are expected to select a motor and corresponding driver that best match their specific chassis and performance requirements.


\section{Software Design}\label{sec:software_desgin}
\begin{figure}[htbp]
  \centering
  \includegraphics{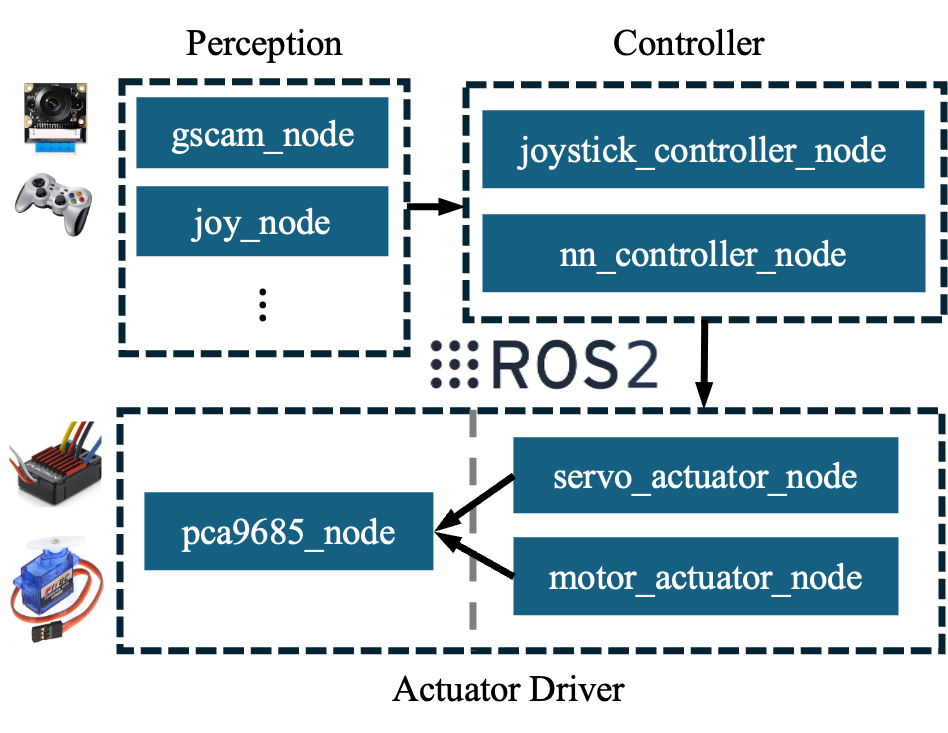}
  \caption{Software system architecture}
  \label{fig:software_system}
\end{figure}

The primary objective of our software design is modularity. To this end, our software stack is based on ROS~2 Humble~\cite{doi:10.1126/scirobotics.abm6074}, which natively supports modular programming by enabling users to split functionality into different nodes and connect them with efficient communication mechanisms, such as topics and services. In addition, the ROS~2 community provides stable packages for joysticks and cameras, saving development time. We split our features into three subsystems, each consisting of one or multiple nodes. The whole software system is shown in Fig.~\ref{fig:software_system}. 

\subsection{Controller}
The controller is responsible for translating user inputs into control messages. This subsystem comprises three nodes: a \textit{joystick\_controller\_node} for manual control and an \textit{nn\_controller\_node} that employs a trained neural network for autonomous driving. A base class \texttt{Controller} is provided for nodes within this subsystem. Child nodes invoke functions implemented by the base class, supplying the target steering angle and/or throttle command. The base class manages message generation and the publishing process.

The message published by the controller is a customized type \textit{motion\_cmd} that has the following three fields:
\begin{enumerate}
    \item Header: Contains timestamp for the recorder to synchronize messages.
    \item Source: The node that sent the controller message.
    \item Value: The target value.
\end{enumerate}

\subsection{Actuator Driver}
The actuator driver is organized into a two-layer architecture. The first layer receives control messages and translates them into PWM pulse widths suitable for driving PWM-based actuators. The current implementation includes two nodes in this layer: \textit{servo\_actuator\_node}, which is responsible for steering control, and \textit{motor\_actuator\_node}, which manages throttle control. All nodes in this layer inherit from a common base class, \texttt{Actuator}. The base class subscribes to control messages published by multiple controller nodes and aggregates the received commands through summation. It then invokes functions implemented by the derived classes, passing the aggregated control command as an argument. Each child node converts this command into a corresponding PWM pulse width and publishes the computed value to the second layer via a ROS~2 message interface.

The second layer translates the received PWM pulse widths into hardware-level actuation signals. This layer currently consists of a single node, \textit{actuator\_node}, which interfaces with the PCA9685 board through the I2C bus. The node configures the PWM frequency and writes the specified pulse width values to the appropriate output channels, thereby generating precise timing signals to drive the actuators.

\subsection{Perception}\label{subsec:perception}
The perception subsystem contains ROS~2 drivers responsible for acquiring sensor measurements and publishing the data using the corresponding ROS~2 sensor message types. For instance, the \textit{gscam\_node} is responsible for acquiring camera images from the onboard sensor and publishing them as ROS~2 \texttt{sensor\_msgs/msg/Image} messages. Similarly, the \textit{joy\_node} converts user inputs from a physical joystick into standardized ROS~2 \texttt{sensor\_msgs/msg/Joy} messages, enabling downstream nodes to process control commands in a unified format.


\section{Evaluation}\label{sec:evaluation}
To evaluate the platform’s suitability for research use, we built a prototype based on the configuration described in Table~\ref{tab:exp_setup} (the design files are available in the \href{https://github.com/Trustworthy-Engineered-Autonomy-Lab/TEACar-Open-Source-Autonomous-Driving-Platform.git}{project repository}\footnote{https://github.com/Trustworthy-Engineered-Autonomy-Lab/TEACar-Open-Source-Autonomous-Driving-Platform.git}). The approximate cost of buying all necessary components and materials was 1200~USD in the United States, circa late February 2026. Compared with RoboRacer, which requires approximately 3000~USD to acquire, the proposed platform significantly reduces the financial barrier to entry while maintaining comparable research functionality. The rest of this section explains our measurements and experiments spanning the mechanical, hardware, and software aspects.

\subsection{Mechanical Evaluation}
Experiments are conducted to measure the car's mechanical characteristics and structural rigidity.

\noindent
\textbf{Mechanical Characteristics}. The mechanical characteristics were measured, and the results are shown in Table~\ref{tab:mech_chara}. The relatively low values highlight a significant advantage of \TEACar in compactness over RoboRacer.

\begin{table}[h]
    \centering
    \caption{Mechanical Characteristics of \TEACar~and RoboRacer.}
    \begin{tabular}{ccc}
        \hline
        \textbf{Name} & \textbf{\TEACar} & \textbf{RoboRacer} \\ \hline
        Total Mass & 1.25 kg & 3.13 kg \\
        Total Length & 285 mm & 555 mm\\
        Total Width & 235 mm & 290 mm   \\
        Total Height & 225 mm & 175 mm\\
        Min Turning Radius & 550 mm & 1200 mm \\
        Max Roll Angle & $38^\circ$ & $43^\circ$  \\
        Max Pitch Angle & $49^\circ$ & $55^\circ$\\ \hline
    \end{tabular}
    \label{tab:mech_chara}
\end{table}



\noindent
\textbf{Structural Rigidity}. Structural rigidity was evaluated through repeatable drop tests. The assembled vehicle was lifted to a height of 1 m and released to impact the ground under identical conditions. This procedure was repeated three times. No structural damage, component detachment, or functional degradation was observed after the tests, indicating sufficient mechanical robustness for experimental use.




\subsection{Hardware and Software Evaluation}
Experiments were designed to assess hardware stability and to check that the software performance is sufficient for research and educational purposes. To this end, hardware and software performance were jointly evaluated by testing three different CNN-based neural network controllers on the prototype platform. Each controller received a single image as input and produced a steering control command as output. For each controller, we measured the system operating time, the average inference latency, and the average power consumption. The camera frame rate was set to 30~FPS, capturing RGB images at a resolution 
of $144 \times 224$ throughout the experiments.

\begin{figure}[htbp]
    \centering
    \vspace{-3mm}
    \includegraphics[width=\linewidth]{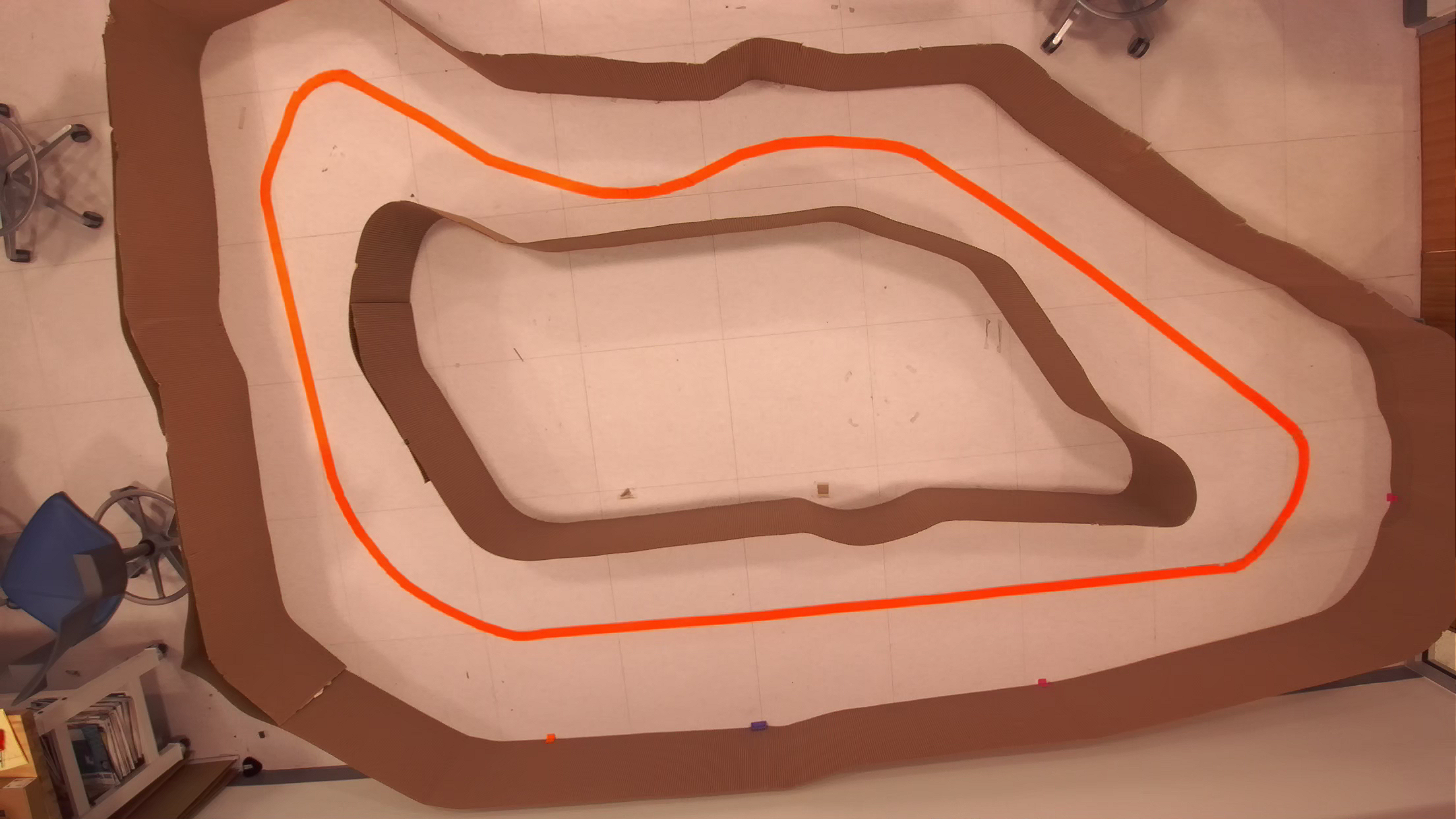}
    \caption{The track for \TEACar~experimentation.}
    \label{fig:track}
    \vspace{-3mm}
\end{figure}

\begin{table}[h]
\caption{Prototype \TEACar~Configuration.}
\label{tab:exp_setup}
\centering
\begin{tabular}{cc}
\hline
\textbf{Part} &  \textbf{Brand \& Model} \\
\hline
Computing Unit & Jeston Orin NX\\
Battery & TRAXXAS 3S LiPo 2300MAH\\
Camera & IMX219-160\\
IMU & MPU6050 \\
Joystick & Logitech Wireless Gamepad F710 \\
Chassis & 1/14-scale RC Chassis \\
\hline
\end{tabular}
\end{table}

\smallskip
\noindent
\textbf{Racing Track.}
We built a track shown in Fig.~\ref{fig:track}. The track was constructed with boundary walls and an orange guideline positioned along its center. 

\smallskip
\noindent
\textbf{Data Collection}.
The human driver raced the car on the track counterclockwise, following the orange line. During operation, the recorder subsystem described in Section~\ref{subsec:perception} was used to collect 10,000 synchronized pairs of camera images and steering angle measurements.

\looseness=-1
\smallskip
\noindent
\textbf{Neural Controller Training}.
Using the collected data, we trained three CNN-based controllers (``small'', ``medium'', ``large'') sharing the same architectural design but differing in parameter counts. The input resolution was set to match the camera image resolution, namely $3 \times 144 \times 224$. Following the convolutional layers, the network generated a feature map. The feature map was flattened and subsequently processed by four fully connected layers to produce a single output value, interpreted as the steering angle of the car. The network architectures and per-layer parameter sizes are summarized in Table~\ref{tab:cnn_arch}, along with the total number of parameters for each controller.

\looseness=-1
\smallskip
\noindent
\textbf{Test Controllers}. 
The trained controllers were deployed on the vehicle, where each controller generated steering commands. During operation, the average power consumption by the computing unit and the average forward-pass inference latency were also measured. For each task, the vehicle was operated continuously for 30~minutes. The initial and final battery voltages were recorded to evaluate energy consumption. The experimental results are summarized in Table~\ref{tab:exp_results}.

The measured average latency of a couple of milliseconds is low, well within an acceptable real-time range for onboard autonomous control. In addition, the average power consumption of the computing unit remained below its rated maximum power budget 25W, indicating that the platform can sustain prolonged operation without exceeding hardware limits.

To compare performance with DonkeyCar, we deployed the three models on its platform and measured inference latency. The results are shown in Table~\ref{tab:comp_latency}. All models run roughly 6 times slower than on our prototype, likely due to the limited compute capability of the onboard \textit{Jetson Nano\texttrademark}.

\begin{table}[h]
\caption{Hardware and Software Experiment Results.}
\label{tab:exp_results}
\centering
\begin{tabular}{cccccc}
\hline
\textbf{CNN} & \multicolumn{2}{c}{\textbf{Battery Voltage (V)}}  &  \textbf{Average} & \textbf{Average} \\ \cline{2-3}
\textbf{Controller} & \textbf{Initial} & \textbf{Final} (+30 min) & \textbf{Latency (ms)} & \textbf{Power (W)} \\
\hline
Small & 12.6 & 11.6 & 1.8 & 7.5\\
Medium & 12.6 & 11.6 & 1.9 & 7.5\\
Large & 12.6 & 11.5 & 2.8 & 7.6\\
\hline
\end{tabular}
\end{table}

\begin{table}[h]
\caption{Comparison of CNN Inference Latency (ms)}
\label{tab:comp_latency}
\centering
\begin{tabular}{ccc}
\hline
\textbf{CNN} & \textbf{DonkeyCar} & \textbf{\TEACar} \\
\textbf{Controller} & & \\
\hline
Small & 12.4 & 1.8\\
Medium & 14.7 & 1.9\\
Large & 16.7 & 2.8\\
\hline
\end{tabular}
\end{table}

\begin{table}[htbp]
\centering
\caption{CNN Architectures for the Driving Controller.}
\label{tab:cnn_arch}
\setlength{\tabcolsep}{4pt}
\begin{tabular}{lccc}
\hline
\textbf{Layer} & \textbf{Small} & \textbf{Medium} & \textbf{Large} \\
\hline\hline
Conv1 & $3\cdot24\cdot5\cdot5$  & $3\cdot24\cdot5\cdot5$ & $3\cdot36\cdot5\cdot5$ \\
(stride=2) & $+24$ & $+24$ & $+36$ \\ \hline
Relu & - & - & -\\ \hline
Conv2 & $24\cdot24\cdot5\cdot5$ & $24\cdot36\cdot5\cdot5$ & $36\cdot48\cdot5\cdot5$\\
(stride=2) & $+24$ & $+36$ & $+48$\\ \hline
Relu & - & - & - \\ \hline
Conv3 & $24\cdot36\cdot5\cdot5$  & $36\cdot48\cdot5\cdot5$ & $48\cdot64\cdot5\cdot5$ \\
(stride=2) & $+36$ & $+48$ & $+64$ \\ \hline
Relu & - & - & -\\ \hline
Conv4 & $36\cdot48\cdot3\cdot3$  & $48\cdot64\cdot3\cdot3$ & $64\cdot96\cdot3\cdot3$ \\
(stride=1) & $+48$ & $+64$ & $+96$\\ \hline
Relu & - & - & - \\ \hline
Conv5 & $48\cdot48\cdot3\cdot3$  & $64\cdot64\cdot3\cdot3$ & $96\cdot96\cdot3\cdot3$ \\
(stride=1) & $+48$ & $+64$ & $+96$ \\ \hline
Relu & - & - & - \\ \hline
FC1 & $11,088\cdot64$ & $14,784\cdot100$ & $22,176\cdot200$ \\
& $+64$ & $+100$ & $+200$ \\ \hline
Relu & - & - & - \\ \hline
FC2 & $64\cdot32$  & $100\cdot50$ & $200\cdot100$ \\
& $+32$ & $+50$ & $+100$ \\ \hline
Relu & - & - & -\\ \hline
FC3 & $32\cdot8$  & $50\cdot10$  & $100\cdot20$ \\
& $+8$ & $+10$ & $+20$ \\ \hline
Relu & - & - & -\\ \hline
FC4 & $8\cdot1+1$   & $10\cdot1+1$  & $20\cdot1+1$ \\ \hline
Tanh & - & - & - \\ \hline
\hline
\textbf{Total} & 786,317 & 1,680,775 & 4,856,829\\  
\textbf{Parameters} & & & \\
\hline
\end{tabular}
\end{table}

\section{Conclusion and Future Work}\label{sec:future_work}


This paper proposes \TEACar, an open and modular autonomous racing platform for research and education. The primary system design goal is modularity across mechanical structure, hardware architecture, and software stack, enabling flexible reconfiguration and rapid experimentation. Experimental evaluation demonstrates that the platform provides stable mechanical performance and sufficient computational capability to support real-time autonomous control tasks.

\looseness=-1
Although the current implementation adopts a modular design, it is presently configured primarily for vision-based experiments. Future extensions will incorporate additional sensing modalities, such as LiDAR and depth cameras, to enable more advanced perception, mapping, and sensor fusion.

\section*{Acknowledgements}

We would like to thank Alejandro La Serna Silva, Fernando Guillen, Gabriel Munoz, Zepei Sun, Clarence Dagins, and Difei Chen for their help in prototyping early \TEACar.

This work was supported in part by the NSF Grants CNS 2440920, CNS 2513076, and CCF 2403616. Any opinions, findings, conclusions, or recommendations expressed in this material are those of the authors and do not necessarily reflect the views of the National Science Foundation (NSF) or the US Government. 







\bibliographystyle{IEEEtran}
\bibliography{references}

\end{document}